\documentclass[runningheads]{llncs}

\usepackage[mobile]{eccv}

\usepackage{eccvabbrv}
\usepackage{graphicx}
\usepackage{booktabs}
\usepackage{algorithm}
\usepackage{algorithmic}
\usepackage{booktabs}
\usepackage{colortbl}  %彩色表格需要加载的宏包
\usepackage{xcolor}
\usepackage{multirow} 
\usepackage{pifont}
\usepackage{amsmath}
\usepackage{amsfonts}
\usepackage{amssymb}
\usepackage{float}
\usepackage{enumitem}
\usepackage[pagebackref,breaklinks,colorlinks,citecolor=eccvblue]{hyperref}
\definecolor{mygray}{gray}{.9}
\definecolor{aliceblue}{rgb}{0.94, 0.97, 1.0}
\definecolor{deeppink}{RGB}{255,20,147}
\definecolor{mygray2}{gray}{.6}
\usepackage{orcidlink}
\usepackage{times}
\newcommand{\tablestyle}[2]{\setlength{\tabcolsep}{#1}\renewcommand{\arraystretch}{#2}\centering\footnotesize}
\DeclareRobustCommand\onedot{\futurelet\@let@token\@onedot}

\usepackage[accsupp]{axessibility}  % Improves PDF readability for those with disabilities.

\begin{document}

\title{EVA: Zero-shot Accurate Attributes and Multi-Object Video Editing} 
\titlerunning{\emph{EVA}}


\author{Xiangpeng Yang\inst{1} \and
Linchao Zhu\inst{2} \and Hehe Fan\inst{2} \and Yi Yang\inst{2}}
\authorrunning{Yang et al.}

\institute{ReLER Lab, University of Technology Sydney \and Zhejiang University \\ 
\url{https://knightyxp.github.io/EVA/}
}

\maketitle

\vspace{-10mm}
\begin{figure*}[h]
  \centering
  \includegraphics[width=\linewidth]{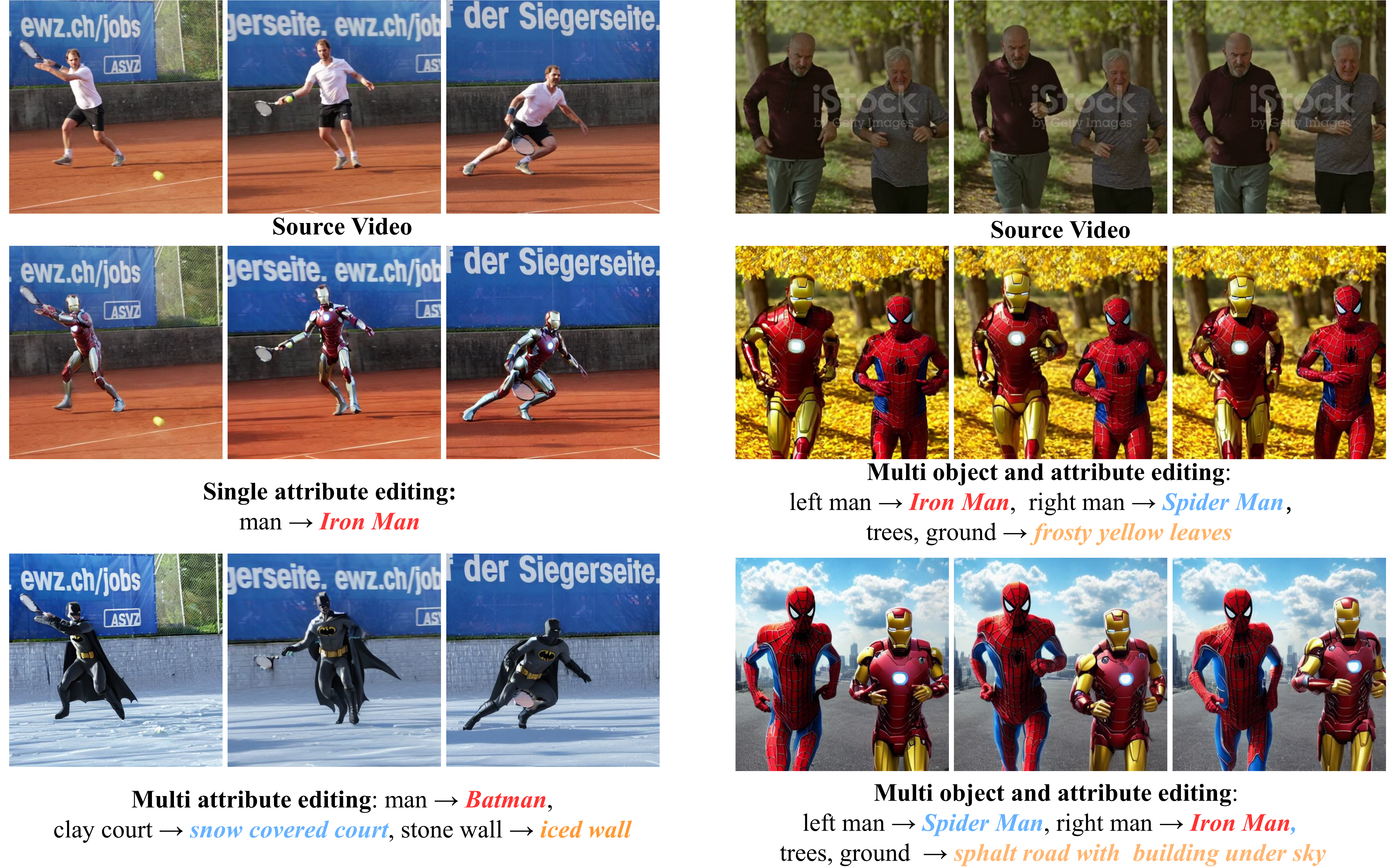}
  \vspace{-6mm}
  \caption{EVA achieves multi-attribute editing for both single and multi-object scenarios, adhering to the source video's layout and faithfully preserving motion information.}
  \vspace{-4mm}
  \label {intro}
\end{figure*}

\vspace{-10mm}
\begin{abstract}
Current diffusion-based video editing primarily focuses on local editing (\textit{e.g.,} object/background editing) or global style editing by utilizing various dense correspondences. However, these methods often fail to accurately edit the foreground and background simultaneously while preserving the original layout.
We find that the crux of the issue stems from the imprecise distribution of attention weights across designated regions, including inaccurate text-to-attribute control and attention leakage.
To tackle this issue, we introduce EVA, a \textbf{zero-shot} and \textbf{multi-attribute} video editing framework tailored for human-centric videos with complex motions.
We incorporate a Spatial-Temporal Layout-Guided Attention mechanism that leverages the intrinsic positive and negative correspondences of cross-frame diffusion features.
To avoid attention leakage, we utilize these correspondences to boost the attention scores of tokens within the same attribute across all video frames while limiting interactions between tokens of different attributes in the self-attention layer.
For precise text-to-attribute manipulation, we use discrete text embeddings focused on specific layout areas within the cross-attention layer.
Benefiting from the precise attention weight distribution, EVA can be easily generalized to multi-object editing scenarios and achieves accurate identity mapping.
Extensive experiments demonstrate EVA achieves state-of-the-art results in real-world scenarios.
Full results are provided at 
\href{https://knightyxp.github.io/EVA/}{project page}.
\vspace{-3mm}
\end{abstract}

% we find the key to simultaneously editing the object and background is the precise distribution of attention weights across designated areas.  

\section{Introduction}
\begin{figure*}[h]
  \vspace{-9mm}
  \centering
  \includegraphics[width=\linewidth]{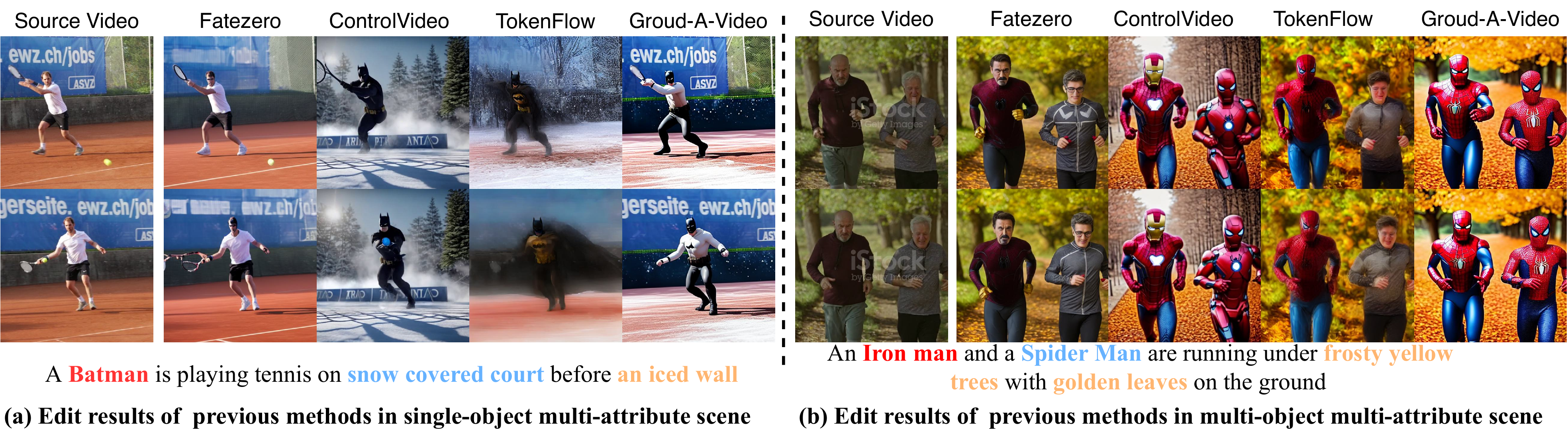}
  \vspace{-7mm}
  \caption{Previous methods failed results are displayed in single/multi-object scenes. 
  EVA’s successful edit result is shown in the third row of Fig \ref{intro} left and the second row of Fig \ref{intro} right.}
  \vspace{-8mm}
  \label {contrast}
\end{figure*}

%%attention leakage到底是什么样的？ 我们的方法为啥可以解决attention leakage?

Text-to-video (T2V) editing, which aims to change the visual appearance of a video according to a given prompt, is an emerging field that harnesses strong generation ability from text-to-image/video models \cite{rombach2022high,ramesh2022hierarchical,saharia2022photorealistic,singer2022make}. 
Previous works have employed dense correspondences, such as depth/edge maps \cite{zhang2023controlvideo, zhao2023controlvideo}, optical flow \cite{yang2023rerender,hu2023videocontrolnet, cong2023flatten} and attention maps \cite{qi2023fatezero}, for local attribute or global style editing, often compromising fidelity.

In this paper, we focus on multi-attribute editing because it enables us to finely adjust local attributes while maintaining the original video's layout and background intact, resulting in more authentic edits.
Previous works have encountered many challenges in multi-attribute editing (Fig \ref{contrast}). The main issues include: (1) overlooking or distorting edits of individual attributes, with FateZero \cite{qi2023fatezero} unable to edit the object and ControlVideo \cite{zhang2023controlvideo} failing  to preserve the background unchanged (Fig \ref{contrast} (a)); (2) the mixing of different attributes, where ControlVideo leads to the blending of textures between "Iron Man" and "Spider-Man", and TokenFlow \cite{geyer2023tokenflow} incorrectly associates the identities of the two characters (Fig \ref{contrast} (b)). 

Ground-A-Video \cite{jeong2023ground} is a recent approach to multi-attribute editing, which employs a cross-frame gated attention mechanism with word-to-bounding box control. Yet, it still has the aforementioned limitations.
The word-to-bounding box control lacks the necessary precision, leading to the loss of fine-grained details, such as the racket illustrated in Fig \ref{contrast} (a).
Furthermore, when bounding boxes are overlapping, it leads to the mixing of textures in adjacent areas.

We identify the imprecise distribution of attention weights as the root cause of these challenges, including inaccurate text-to-attribute control and attention leakage.
% We identify the imprecise distribution of attention weights, which primarily stems from inaccurate text-to-attribute control and attention leakage.
To ensure precise attention weight distribution, we introduce a Spatial-Temporal Layout-Guided Attention (ST-Layout Attn) mechanism. 

First, to achieve accurate text-to-attribute control, we extract the corresponding text embedding for each attribute from the global prompt. These discrete text embeddings are applied to each attribute's corresponding layout areas within the cross-attention layer. For each layout area, we use masks as spatially disentangled information, leveraging the inherent characteristic that masks do not overlap.
To further keep the fine-grained details, we perform latent blend \cite{avrahami2023blended} to preserve the undesired edit areas. 

% Second, to avoid attention leakage, we enhance awareness of negative examples between different attributes across frames.

\begin{figure}[t]
  \centering
   \vspace{-5mm}
  \includegraphics[height=3cm]{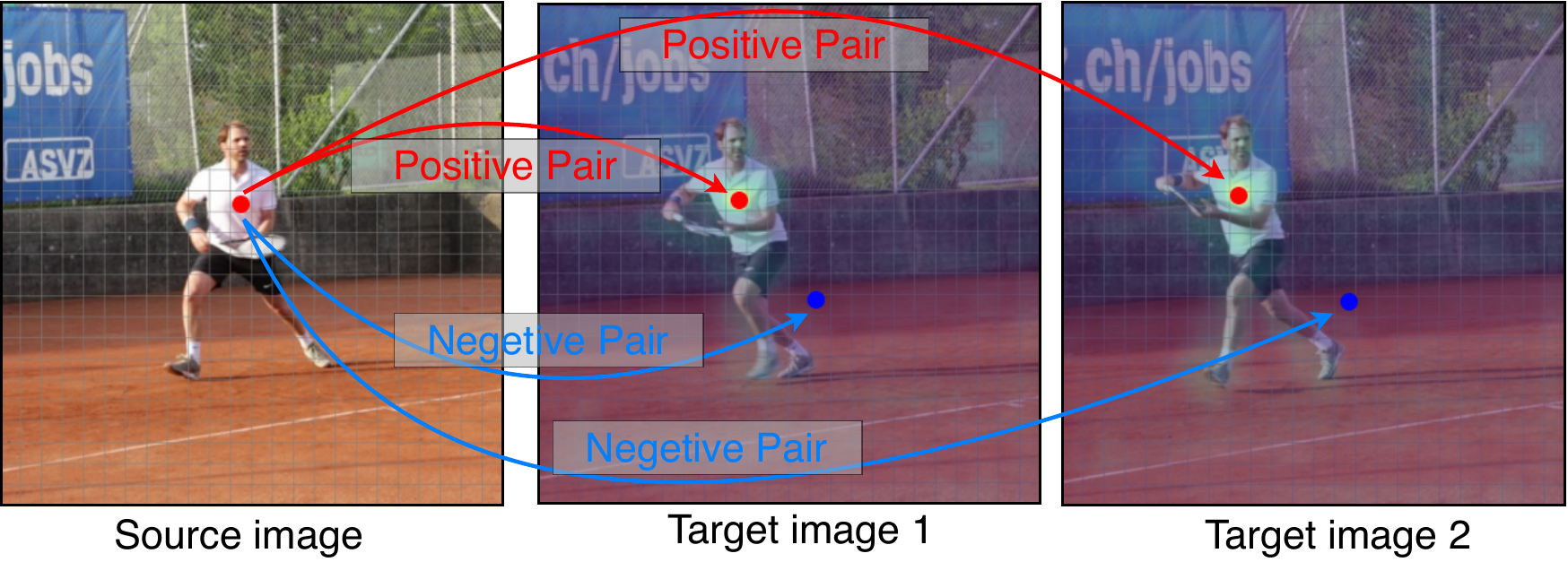}
  \vspace{-2mm}
  \caption{Intrinsic Cross-frame DIFT \cite{tang2023emergent} feature correspondence. We randomly select a ``red point" in the source image, extract its DIFT feature, and compute cosine similarity with the target image. The target's "red point" marks the highest similarity, and "blue point" is the lowest, showing the potential to unsupervised identify intra/inter attributes correspondence.
 }
  \vspace{-7mm}
  \label {dift}
\end{figure}

Second, to avoid attention leakage, our goal is to ensure the mutual exclusivity of different attributes while enhancing the correlation of the same attribute across frames. We leverage the cross-frame diffusion feature similarity (DIFT\cite{tang2023emergent}), which reveals the intrinsic correlations across inter/intra attributes along a spatial-temporal axis. As demonstrated in Fig \ref{dift}, for each token, we identify its corresponding positive pair in other frames (sharing the same attribute) by maximizing the cosine DIFT similarity. Similarly, we determine its negative pair (across different attributes) by minimizing this similarity. 
Thus, leveraging this intrinsic correspondence, we assign positive and negative values to each token across various layouts on a spatial-temporal axis.
Consequently, we enhance the attention scores for tokens within the same attribute and limit interactions between tokens of different attributes across frames, thus significantly mitigating attention leakage.

Benefiting from precise attention-weight distribution and text-to-attribute control brought by our attention mechanism, we realize accurate identity mapping and background editing in the multi-object scenario. Such as swapping identity while editing the background (Fig \ref{intro} right). 

Our key contributions can be summarized as follows:
\begin{itemize}
\item We propose EVA, a general framework for accurate attributes and multi-object video editing, which realizes accurate weight distribution and identity mapping.
\item Leveraging intrinsic cross-frame DIFT correspondence, we introduce ST-Layout Attn for accurate text-to-attribute control and to avoid attention leakage.
\item Without tuning any parameters, we achieve state-of-the-art results on existing benchmarks and real-world videos both qualitatively and quantitatively.
\end{itemize}

\section{Related Work}

\subsection{Text-to-Image Editing/Generation}

In the realm of single attribute text-to-image editing, various approaches have been explored, from manipulating attention maps in Pix2Pix-Zero \cite{parmar2023zero} and Prompt2Prompt \cite{hertz2022prompt} to employing masks in DiffEdit \cite{couairon2022diffedit} and Latent Blend \cite{avrahami2022blended,avrahami2023blended} for foreground modifications while preserving the background. 

For multi-attribute editing, efforts such as Attention and Excite \cite{chefer2023attend} and DPL \cite{wang2023dynamic} focus on maximizing the attention scores for each subject token and reducing attention leakage. 
Recently, in single image generation, \cite{densediffusion} adjusted modulate attention value according to layout masks and dense caption. \cite{phung2023grounded} proposed attention refocus loss for regularization. 
However, guiding the attention map solely with single-frame layout masks and dense captioning is inadequate in video editing, as it fails to maintain the original video’s integrity and temporal consistency.
% Another trend in image generation is \cite{ge2023expressive} and \cite{densediffusion}, applying region-specific prompts and modulating attention value for enhanced layout control. Despite these advances, accurately editing multiple attributes in videos while maintaining the original layout remains challenging, prompting us to adopt and extend mask-based methods for video editing.

\subsection{Text-to-Video Editing}
\textbf{Video Editing based on Image Diffusion Models}
Tune-A-Video (TAV) \cite{wu2022tune} is the first work to extend latent diffusion models to the spatial-temporal domain and encode the source motion implicitly by one-shot tuning but still fails to preserve local details. Fatezero \cite{qi2023fatezero} is a prompt2prompt \cite{hertz2022prompt} based editing method, fusing self- and cross-attention maps for temporal consistency. 
However, it requires extensive RAM usage and suffers from layout preservation even when equipping TAV for local object editing. 
%up to 200-300GB for a 16-frame video, due to maintaining each frame's inversion 
\cite{chai2023stablevideo} and \cite{ouyang2023codef}, following the Neural Atlas \cite{kasten2021layered} or dynamic Nerf's deformation field \cite{mildenhall2021nerf,pumarola2021d}, struggle with non-grid human motion. 
Subsequent methods like Rerender-A-Video \cite{yang2023rerender}, Flatten \cite{cong2023flatten} ControlVideo \cite{zhao2023controlvideo,zhang2023controlvideo} achieve strict temporal consistency via optical-flow, depth/edge maps, but failed in multi-attribute editing while preserving original layouts. Tokenflow \cite{geyer2023tokenflow} enforces a linear mix of nearest key-frame features to ensure consistency but results in detail loss. Ground-A-VIDEO \cite{jeong2023ground} leverages groundings for multi-attribute editing, but it suffers from attention leakage when bounding boxes overlap, even with dense guidance such as optical flow.

\noindent\textbf{Video Editing based on Video Diffusion Models}
% Using text-to-video generation foundation models ~\citep{blattmann2023align, yu2023magvit,} for video editing is a straightforward solution. However, since these models typically require substantial computational resources and gigantic amounts of video data, most models are not open-sourced. As a result, 
Previous video editing work primarily utilized text-to-image SD model \cite{rombach2022high}. 
Recent advancements in video foundation models \cite{blattmann2023align, yu2023magvit, guo2023animatediff, wang2023modelscope} have led efforts like MotionDirector \cite{zhao2023motiondirector} and VideoSwap \cite{gu2023videoswap}  to employ temporal priors for customized motion transfer.
 Yet, current video foundation models are limited to fixed views and struggle with complex human motions. Additionally, these editing methods require tuning parameters, which poses a challenge for real-time video editing applications. In contrast, our EVA method requires no parameter tuning, enabling zero-shot, multi-object and multi-attribute video editing.

\vspace{-5mm}
\section{EVA}
\vspace{-3mm}
In this section, we start by analyzing the prerequisite for multi-attribute editing and the necessity for precise attention weight distribution in section \ref{sec: motivation}.
Subsequently, we present an overview of the proposed EVA pipeline in section \ref{sec: eva}. Following that, we detail the Spatial-Temporal Layout Attention mechanism in \ref{sec: st-layout-attn}. Notice that, our EVA is a general zero-shot framework for editing both single and multiple objects, as well as backgrounds, in human-centric videos with complex motion.
\vspace{-5mm}
\subsection{What is the Key to Multi-Attribute Video Editing?}
\label{sec: motivation}
\vspace{-2mm}
% Following \cite{jeong2023ground}, we define multi-attribute editing as the simultaneous modification of several attributes, described by the formula $\Delta\tau=\{\tau_a \rightarrow \tau_{a'}, \tau_b \rightarrow \tau_{b'}, \tau_c, \cdots\}$, where $\tau$ represents distinct attributes, $\Delta\tau$ is the global prompt, and $a$ and $a'$ denote the source and edited attribute captions, respectively.

% \vspace{-5mm}
\subsubsection{Accurate Text-to-Attribute Control}
\begin{figure}[t]
  \centering
  \includegraphics[width=\linewidth]{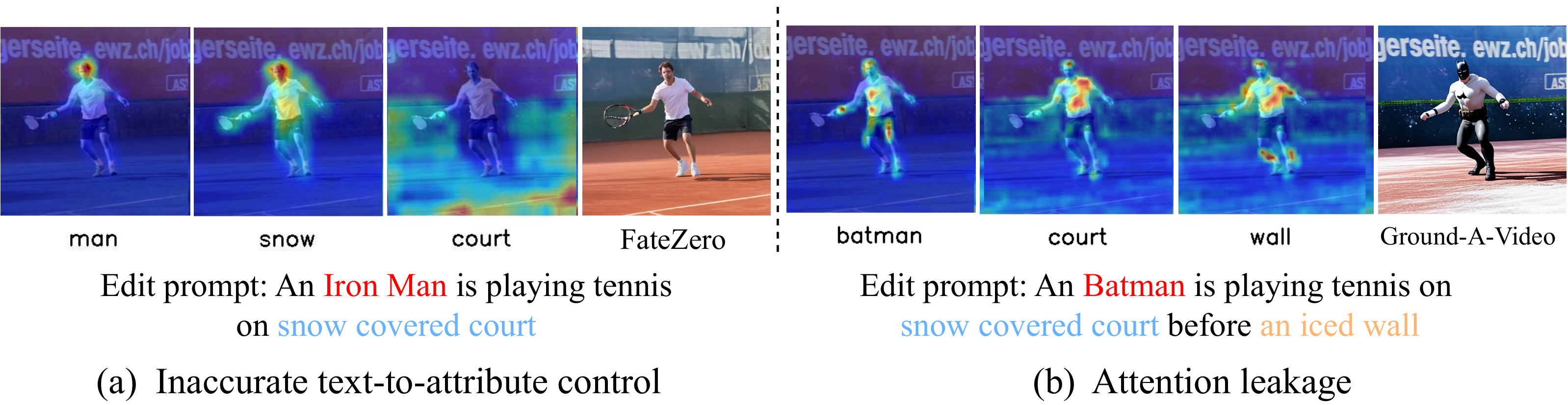}
  \vspace{-5mm}
  \caption{\textit{Left}: FateZero \cite{qi2023fatezero} fails in text-to-attribute control, incorrectly allocating weights to "snow" and not fully covering "man." \textit{Right}: Although Ground-A-Video \cite{jeong2023ground} attempts to ground each attribute individually, it still suffers from attention leakage, leading to texture blending on "Batman's" upper body and imprecise edits of "court" and "wall."}
    \vspace{-4mm}
  \label {motivation}
\end{figure}

% $\tau_\text{man} \rightarrow \tau_\text{Iron Man}$ and $\tau_\text{clay court} \rightarrow \tau_\text{snow covered court}$ 
To accurately edit multiple attributes in a video, it is essential to ensure the model's capability for precise editing of each individual attribute. 
The previous method, FateZero \cite{qi2023fatezero}, addresses this challenge by implementing word-swap in prompt-to-prompt editing \cite{hertz2022prompt}, making the editing of multiple attributes simultaneously. It also merges attention maps from the inversion process to preserve the original motion and layout information. However, as depicted in Fig \ref{motivation} (a), FateZero struggles to edit ``man" to ``Iron Man" and ``clay court" to ``snow covered court" in video with complex motion, even with pose guidance from ControlNet \cite{Zhang_2023_ICCV}.
This issue stems from the misalignment of attribute weights with their appropriate spatial regions. Specifically, the weight of ``man" is concentrated only on the head, and the weight of "snow" is wrongly assigned to "man."  Additionally, "court"'s weight doesn't correctly cover its area, spreading around the person instead.

This challenge inspires us to employ spatially disentangled semantic masks as layout information, leveraging their natural ability to capture each attribute's shape features. Moreover, we aim to ensure that each text embedding exclusively focuses on its corresponding attribute's spatial region, thus ensuring accurate text-to-attribute control.
% We acquire the intended editing area masks by point-based user-interactive SAM-Track \cite{cheng2023segment}. PCA {\&} clustering or thresholding from cross-attention maps falls short in accurately isolating tiny objects like ``tennis ball" and ``racket".

\vspace{-5mm}
\subsubsection{Avoiding Attention Leakage}
However, relying solely on spatially disentangled information is insufficient. 
Previous work Ground-A-Video \cite{jeong2023ground} introduced gated attention from \cite{li2023gligen} for text-to-bounding box control in multi-attribute editing. Yet, in complex motion, \cite{jeong2023ground} still faces challenges in accurately editing the man, ground, and walls.

% As illustrated in  \ref{motivation} (b), our visualization of the cross-attention map reveals that weights from other attributes, such as ``court" and ``wall," leak onto the ``Batman," causing his upper body to appear white.
As illustrated in Fig \ref{motivation} (b), our visualization of the cross-attention map reveals a phenomenon termed "attention leakage," where weights from other attributes, such as "court" and "wall," leak onto "Batman," causing his upper body to appear white.
To address this, we introduce negative example awareness among different attributes to ensure the mutual exclusivity of each attribute weight, thereby avoiding attention leakage. 
Additionally, the incomplete weight distribution across Batman’s body underscores the necessity to strengthen the correlation within each attribute.
Cross-frame DIFT features inherently exist this kind of correspondence. For a query token, cross-frame DIFT feature similarity identifies positive pairs within the same attribute across frames and negative pairs in different attributes by calculating max/min similarity, as shown in Fig \ref{dift}.
Leveraging this correspondence, we introduce ST-Layout Attn mechanism. 
Our ST-Layout Attn not only ensures that each text embedding concentrates on its respective attribute, but also enhances the internal coherence within attributes and keeps the exclusivity of attention weights among different attributes. Through this approach, we effectively achieve accurate text-attribute control and prevent attention leakage.

%$\tau_{man} \rightarrow \tau_{Batman}, \tau_{clay\, court} \rightarrow \tau_{snow\,covered\, court}, \tau_{stone \, wall} \rightarrow \tau_{iced\, wall} $.

% Cross-frame DIFT features naturally exhibit this kind of negative correspondence between different attributes, as illustrated in Fig \ref{dift}. 
\vspace{-2mm}
\subsection{Overall Framework}
\label{sec: eva}
%% 结合图中的notation去讲
Our framework aims to edit the source video $V^{1:N}$ according to a textual prompt $\Delta_\tau$ which contains a series of desired local attribute edits $\{\tau_{1}{\rightarrow}\tau_{1'},\tau_{2}{\rightarrow}\tau_{2'},\cdots \}$. Following previous work \cite{wu2022tune,qi2023fatezero}, we inflated the original StableDiffusion\cite{rombach2022high} (SD) along the temporal axis to adapt for 3-dimension video input.

For human-centric complex motion, we want to decouple human motion from object identity. Thus, we directly utilize the human pose as sparse motion information from the source video object. Following  \cite{zhao2023controlvideo,zhang2023controlvideo}, we employ ControlNet's \cite{Zhang_2023_ICCV} pose guidance to promote temporal consistency.

\begin{figure*}[t]
  \centering
  % \vspace{-mm}
  \includegraphics[width=\linewidth]{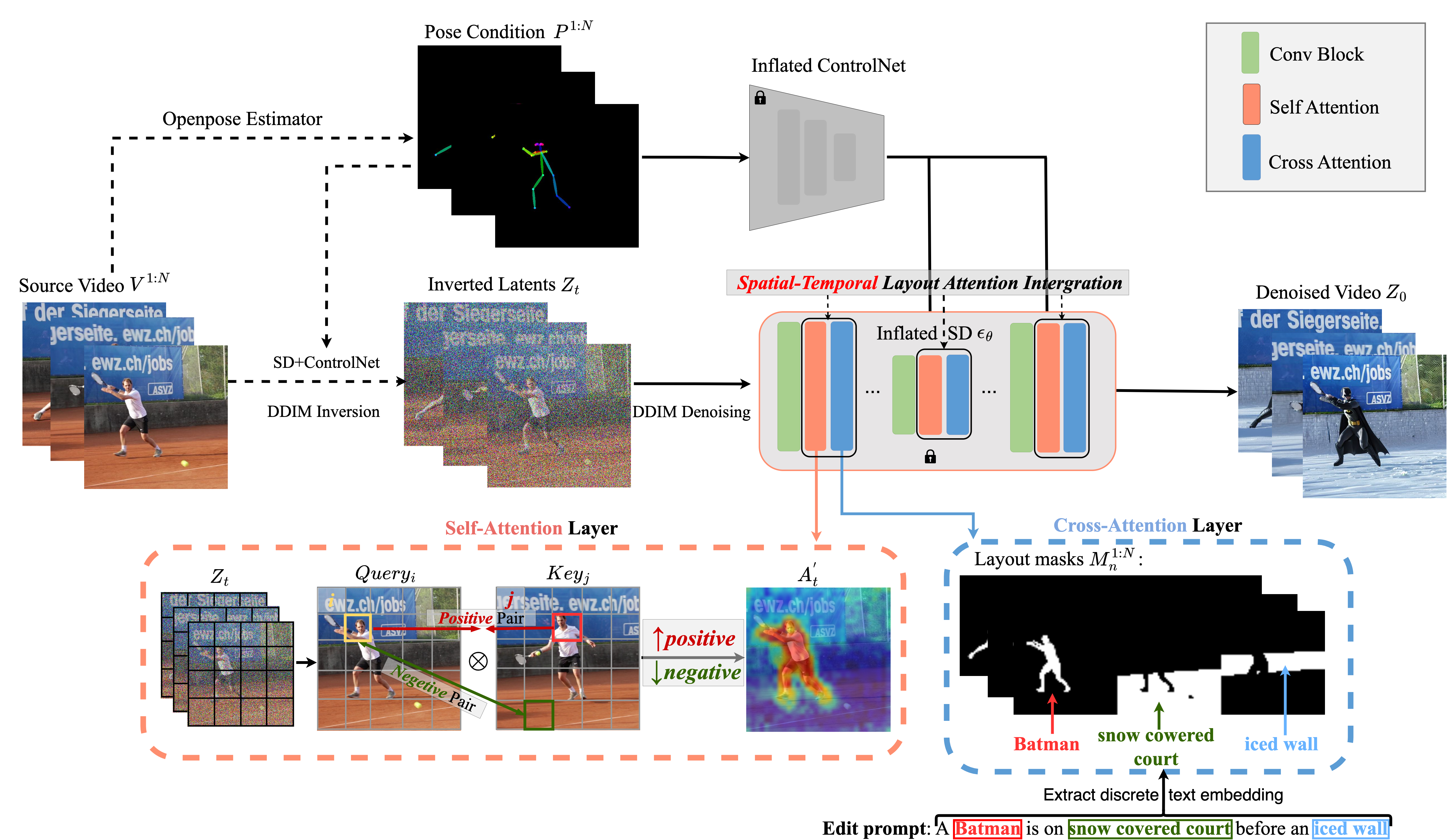}
  \vspace{-5mm}
  \caption{EVA pipeline. We integrate the ST-Layout Attn within the frozen SD in the denoising process.
  In the self-attention layer, 
  we compute the positive/negative value of each query token in different attributes from a spatial-temporal perspective, 
  This allows us to augment the attention scores for tokens within the same attribute and reduce them for tokens in different attributes. 
 In the cross-attention layer, we extract each attribute's text embeddings from the edit prompt, ensuring they focus only on corresponding layouts across frames.  
  }
  \vspace{-7mm}
  \label {framework}
\end{figure*}

Fig \ref{framework} illustrates the overall pipeline of our EVA:

(1) Firstly, we obtain layout masks $M_n^{1:N}$ ($n$ denotes the layout or attribute classes) corresponding to each attribute through user-interactive Segment-and-Tracking anything \cite{cheng2023segment}, which provides crucial layout information. 
We obtain pose condition $P^{1:N}$ through the OpenPose estimator \cite{cao1812openpose} to encode source complex motion information. Furthermore, we extract text embeddings for each attribute from the edit prompt, setting the stage for their subsequent use in the cross-attention layer of ST-Layout Attn. 

(2) Then, the input videos undergo DDIM inversion in the latent space of Stable Diffusion (SD \cite{rombach2022high}) and ControlNet \cite{Zhang_2023_ICCV} to enhance the fidelity of the generated video. 

(3) Finally, the inverted Latents ${Z}_t$ are fed into inflated SD and ControlNet during the DDIM denoising process.
In the denoising process, we incorporate ST-Layout Attn to ensure accurate attention weight distribution in a zero-shot manner. 
In the self-attention layer, based on the cross-frame DIFT similarity, we boost the attention scores of tokens in the same attribute and restrict communication between different attributes across all frames, avoiding attention leakage.
In the cross-attention layer, we utilize each attribute's text embedding to enable direct text-to-attribute control.

% %laying the groundwork for precise attention modulation.

% (3) 
% All the steps are implemented in a zero-shot manner.

\subsection{Spatial-Temporal Layout-Guided Attention}
\label{sec: st-layout-attn}
DenseDiffusion \cite{densediffusion} proposed modulating intermediate attention maps according to layout mask guidance for single image generation. However, in the context of dynamic video scenes,  the correlation between attributes across different frames constantly changes.
Therefore, enhancing intra-attribute similarity while reducing inter-attribute interaction becomes crucial in scenes with complex motion.

% DenseDiffusion \cite{densediffusion} proposed to modulate the intermediate attention maps according to the layout mask guidance for single image generation
% In the context of dynamic video scenes, however, the correlation between different attributes across different frames is constantly changing. Addressing the challenge of enhancing intra-attribute similarity while reducing inter-attribute interaction becomes crucial in complex motion scenes.
% Specifically, they consider the value range of original attention scores and adjust the degree according to each dense caption for each attribute.
% We follow this manner but try to expand the receptive field of positive/negative values to accommodate dynamic scenes in videos.
%in the single image generation, they compute the maximum/minimum value for each query in the attention map and increase/decrease each attribute's attention value accordingly. 

\textbf{Notations}
We define a set of layout masks $m_L^i={[}m_1^i, m_2^i,\cdots,m_l^i{]} $ for the $i^{th}$ frame and multi attributes $\tau_L={[}\tau_1, \tau_2,\cdots,\tau_l{]} $ ,where $L$ denotes total classes of layout attributes, and each pair $(\tau_l, m_l)$ correspond to a single region.

\textbf{Identify the correlations of intra/inter attributes}
Consider the original SD is trained on the large-scale images, and lacks a built-in temporal module in the pretraining process. 
To incorporate temporal information effectively, we treat the full video frames as ``a larger picture''. 
Specifically, for each query $Q$ at frame $i$, the key $K$ or value $V$ is computed from the concatenated latents across all frames, this process can be formulated as:
\begin{equation}
Q = W^{Q} {z}^{i}_{t},\quad
K = W^{K} {z}^{N}_{t},\quad
V = W^{V} {z}^{N}_{t},
\end{equation}
where $ W^{Q}$,$ W^{K}$,$ W^{V}$ project $z_t$ into query, key and value.
${z}^{N}_{t} = [{z}^{1}_{t}, \cdots ,{z}^{n}_{t}]$  denotes the concatenation of  each frame latent state and $n$ represents the total video frames. 

Continually, we need to find each attribute's correlations across different frames.
As illustrated in Fig \ref{dift}, the maximum value in cross-frame DIFT feature similarity indicates the strongest response among tokens within the same attribute, whereas the minimum similarity points to the relationship between tokens of different attributes. To discern the relationship of each query token with the same and different attributes throughout the video, we identify the spatial-temporal 
 positive/negative value for each query on the spatial-temporal axis as follows:

\begin{equation}
\begin{gathered}
    M_{\text{pos}}^i = \mbox{max}({Q^i} [{K}^{1},\cdots,{K}^{n}]^\top)- {Q^i} {[{K}^{1},\cdots,{K}^{n}]^\top)} \\
    M_{\text{neg}}^i =  {Q^i} {[{K}^{1},\cdots ,{K}^{n}]^\top} - \mbox{min}({Q^i} {[{K}^{1},\cdots,{K}^{n}]}^\top) \\
\end{gathered}
\end{equation}
% where $n$ represents the total video frames. 
These spatial-temporal positive/negative values represent the relationships within the same/different attributes, respectively, allowing us to enhance attention scores among tokens of the same attribute and reduce them among tokens of different attributes to avoid attention leakage.

\textbf{Modulate Spatial-Temporal Attention Value}
We follow \cite{densediffusion}, and modulate the attention map $A_i$ to $A_i'$ for each frame $i$ based on the spatial-temporal positive/negative value, this can be formulated as:
\begin{equation}
\centering
\begin{gathered}
    A_i' = \mbox{softmax}(\frac{Q^i [K^N]^\top + M}{\sqrt{d}}), \\
    M = \lambda_{t} \cdot {R}^i_{\text{st}} \odot M^i_{\text{pos}} \odot (1-S^i_{\text{st}})-\lambda_t \cdot (1-{R}^i_{\text{st}}) \odot M^i_{\text{neg}} \odot (1-S^i_{\text{st}}), 
\end{gathered}
\label{eq:st-layout attn}
\end{equation}
where ${R}^i_{\text{st}} \in \mathbb{R}^{|\text{queries}| \times |\text{keys}|}$ indicates the query-key pair condition map at frame $i$,
manipulating whether to increase or decrease the attention score for a particular pair. 
For the tokens in the same attribute across different frames, which will be viewed as a positive pair, leading to an increase in their attention score. 
In contrast, when the tokens are from different attributes (layouts) in the video, they constitute a negative pair, resulting in a reduced attention score. $\lambda_{t}$ is a regularization parameter for timestep $t$, controlling modulation function intensity. 
$S^i_{\text{st}}$ represents the spatial-temporal regularization for each attribute size. We calculate each attribute class area proportions across video frames, enabling dynamic attention weight adjustments to layout size variations.

% $S_i$ represents the regularization for each object size. In videos, the area of each layout changes dynamically. To automatically adjust the attention weight according to the size of each layout, we extend the original size regularization to a spatial-temporal dimension. This involves calculating the area proportion of each attribute class across all frames in the video, increasing the control degree for smaller areas and decreasing it for larger ones.

\textbf{Regularize Self Attention Map Beyond Spatial}.
In the self-attention layer, we aim to avoid attention leakage by increasing attention scores for tokens within the same attribute, while restricting interactions between tokens in different attributes within the same frame or across various frames. Consequently, our query-key condition map is defined as:
\begin{equation} 
\centering
{R}_{\text{st}}^{(i), \text{self}} := \left\{
    \begin{array}{ll}
        0,  \forall j \in [1:N], \text{if }   {m}_l^{(i)}[a] \neq {m}_l^{(j)}[b], \\
        1,  \text{otherwise} \\
    \end{array}
\right.
\label{eq:am_self_time}
\end{equation}
%$\mR \in \R^{T\times H \times W }$,
where $a$ and $b$ are token indexes of the query and key in the condition map, respectively. $i,j$ are frame indices, and ${m}_l$ represents a binary map for a single attribute. If tokens belong to different attributes across frames, the value is zero.
% Where $a$ and $b$ are the token indexes of the query and key in the condition map, respectively.  $i,j$ are frame indices, ${m}_n$ represents a layout attribute vector. If 、query and key belong to different attributes across frames, the condition map value is zero.

%with $i$ being the current frame and $j$ representing a different frame.

\textbf{Discrete Text control in Cross-Attention Layers}
In the cross-attention layer, to achieve precise text-to-attribute control, we employ discrete text embeddings for each attribute focused on corresponding layouts. Based on layout masks $m^i_L$, we tailor the cross-attention query-key condition maps for textual cues to target specific regions:
\begin{equation}
\centering
\begin{gathered}
    {R}_{\text{st}}^{(i), \text{cross}} := \Bigl\{
        \begin{array}{ll}
        {0}, & \text{if } {k}[b] = 0 \\
        {m}^i_{{k}[b]}, & \text{otherwise}\\
        \end{array} \\
\end{gathered}
\label{eq: cross-attention qk map}
\end{equation}
where ${m}^i_{{k}[b]}$ is a binary map fitting the spatial resolution at $i^{th}$frame. ${k}[b] \in \mathbb{R}^{|\text{keys}|}$ maps the $b_{th}$ text token to its attribute index, with zero indicating no association.
Take the phrase "An Iron Man on a snow covered court": we have two attributes, with $\tau_{1} = \text{man}$ and $\tau_2 = \text{court}$. 
The value of  ${k}[0,3,4]$ is zero for unrelated tokens, ${k}[1,2] = 1$ for "Iron Man", and ${k}[5,6,7] = 2$ for "snow covered court".
% The mapping is ${k}[0] = 0$  and  ${k}[3,4] = 0$ for unrelated tokens, ${k}[1,2] = 1$ for "Iron Man", and ${k}[5,6,7] = 2$ for "snow-covered court".

% In the cross-attention layer, text features interact with intermediate video frame features. However, global text prompts often fail to provide precise word-to-attribute control. Therefore, we want to use discrete target prompts to control each attribute directly. Based on the layout conditions $m^i_N$, we adjust the cross-attention query-key condition maps  to allow specific textual clues to focus within designated regions:

\section{Experiments}

\subsection{Experimental Settings}
\textbf{Datasets}
We validate our EVA model on a dataset comprising 26 videos, sourced from DAVIS \cite{perazzi2016benchmark},
TGVE\footnote{\scriptsize{\url{https://sites.google.com/view/loveucvpr23/track4}}},
and the Internet\footnote{\scriptsize{\url{https://www.istockphoto.com/}}}. 
This dataset includes 14 single-object and 12 multi-object human-centric complex motion videos. For each video, we manually annotate the descriptions of the source video and create 3 creative textual prompts, encompassing single-attribute, multi-attribute, multi-object and background editing. Ultimately, this process results in the construction of 78 video-text pairs. Each video is cropped and resized to a resolution of 512x512, containing 16-32 frames. 

% Considering that human motion typically unfolds over an extended period, we set the frame rate to 1 to capture this dynamic effectively.
\noindent\textbf{Metrics}
Following~\cite{wu2022tune,qi2023fatezero}, we assess the video quality using five metrics:
\textbf{Frame Acc} measures frame-wise editing accuracy, which computes the percentage of frames with higher CLIP similarity to the target prompt than the source, following \cite{qi2023fatezero}.
\textbf{CLIP-T} is the average cosine similarity between the input prompt and all video frames, which is used to measure textual alignment.
We also follow \cite{geyer2023tokenflow,cong2023flatten} to measure the temporal consistency by CLIP-F and Warp-Error\cite{lai2018learning}. 
\textbf{CLIP-F} measures the average cosine similarity between all pairs of consecutive frames, indicating global-level temporal consistency.
\textbf{Warp-Err} calculates the pixel-level difference by warping the edited video frames according to the estimated optical flow of the source video, extracted by RAFT-Large\cite{teed2020raft}. This metric provides a more detailed measure of temporal consistency at the pixel level.
Assessing editing performance solely with these metrics may not offer a holistic view, as unedited videos could still yield low Warp-Err or high CLIP-F scores.
Therefore, following \cite{cong2023flatten}, we adopt $\textbf{Q-edit} = \text{CLIP-T} {/}\text{Wrap-Err}$ as a comprehensive score for video editing quality. For brevity, we scale up Frame Acc/CLIP-F/CLIP-T/Warp-Err all by 100.

\begin{figure*}[t]
  \includegraphics[width=\linewidth]{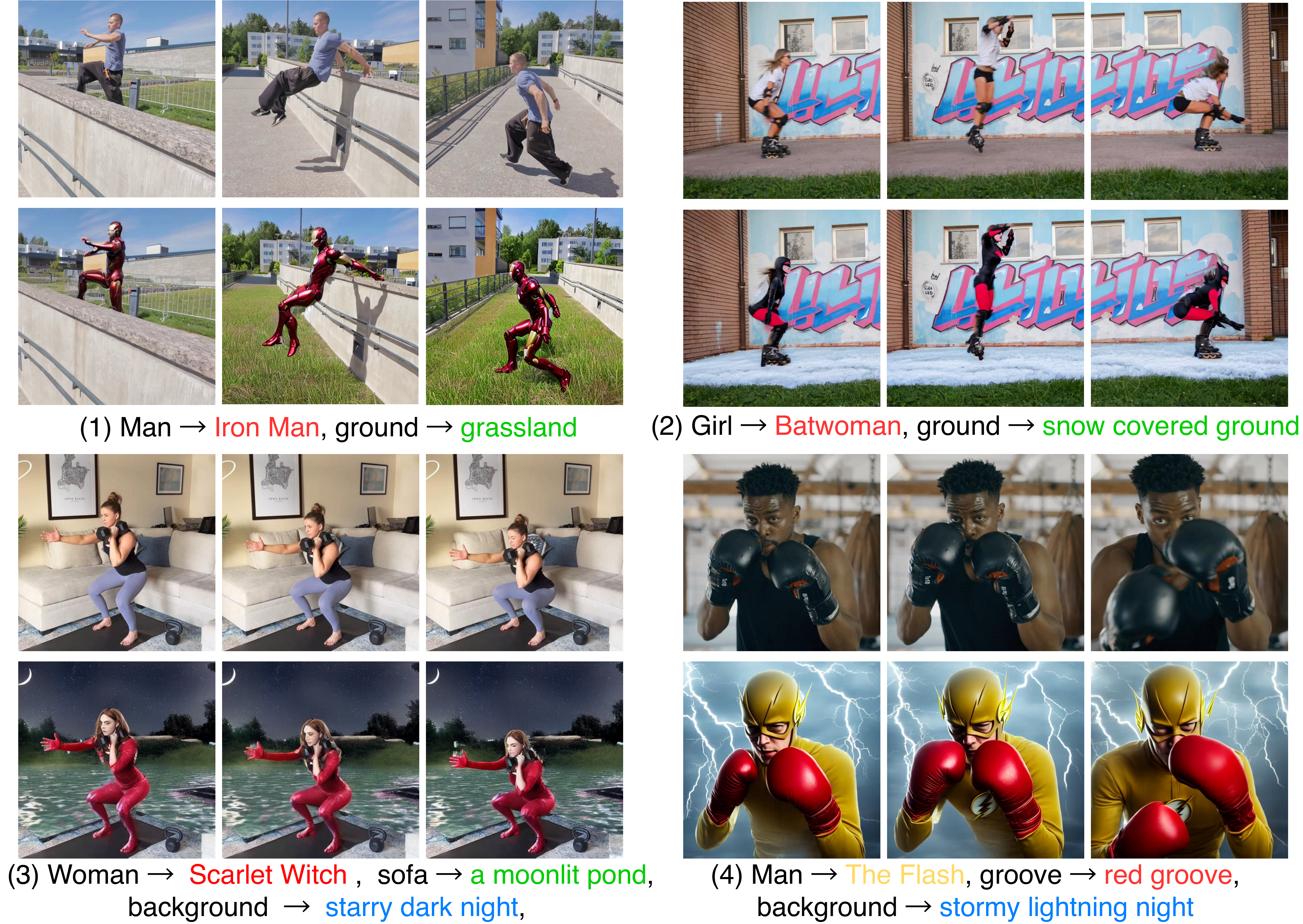}
  \caption{Single-object multi-attribute editing results. We refer the reader to our \href{https://knightyxp.github.io/EVA/}{webpage} for more examples and full-video results. }
  \vspace{-8mm}
  \label {eva-single}
\end{figure*}

\noindent\textbf{Implementation details}
For our implementation, we inflate a pretrained 2D Stable Diffusion \cite{rombach2022high} v1.5 model along with ControlNet \cite{Zhang_2023_ICCV} as the pretrained model. We employ the user-interactive mode of SAM-Track \cite{cheng2023segment} for layout condition, which allows users to specify the areas they wish to edit by clicking to create masks. 
PCA \& clustering or thresholding from cross-attention maps falls short in accurately isolating tiny objects such as ``tennis ball" and ``racket" due to their limited resolution.
To enhance the consistency of edited videos, we adopt DDIM inversion. Our DDIM inversion and denoising steps are all set to 50. 
To improve efficiency, we have implemented slice attention within ST Layout Attn, which further saves memory usage.
We apply ST Layout Attn in the initial 15 denoising steps and set other hyper-parameters the same as \cite{densediffusion}.
All the experiments are done with one NVIDIA A40 GPU.

% \noindent\textbf{Implementation details}
% For our implementation, we inflate a pretrained 2D Stable Diffusion \cite{rombach2022high} v1.5 model along with ControlNet \cite{Zhang_2023_ICCV} as the pretrained model. We employ the user-interactive mode of SAM-Track \cite{cheng2023segment} for layout condition, which allows users to specify the areas they wish to edit by clicking to create masks. 
% PCA\&clustering or thresholding from cross-attention maps falls short in accurately isolating tiny objects such as ``tennis ball" and ``racket" due to their limited resolution.
% To enhance the consistency of edited videos, we adopt DDIM inversion. Our DDIM inversion and denoising \cite{song2020denoising} steps are set at 50. 
% To improve efficiency, we engineer slice attention within ST Layout Attn, which further saves memory usage.
% We apply ST Layout Attn in the inital 15 denoising steps and set other hyper-parameters the same as \cite{densediffusion}.
%  All the experiments are done with one NVIDIA A40 GPU. 

\noindent \textbf{Baselines}
We compare with 4 state-of-the-art video editing methods:
(1) Fatezero \cite{qi2023fatezero} preserves layout information using source video attention maps.
(2) ControlVideo\cite{zhang2023controlvideo} is a training-free method conditioned on ControlNet \cite{Zhang_2023_ICCV}.
(3) Tokenflow \cite{geyer2023tokenflow} samples keyframes and performs linear combinations of features for visual consistency.
(4) GroundVideo \cite{jeong2023ground} uses a word-to-bounding box approach for multi-attribute control. For fairness, all baselines are equipped with ControlNet pose guidance. 
\vspace{-5mm}

\begin{figure*}[t]
  \includegraphics[width=\linewidth]{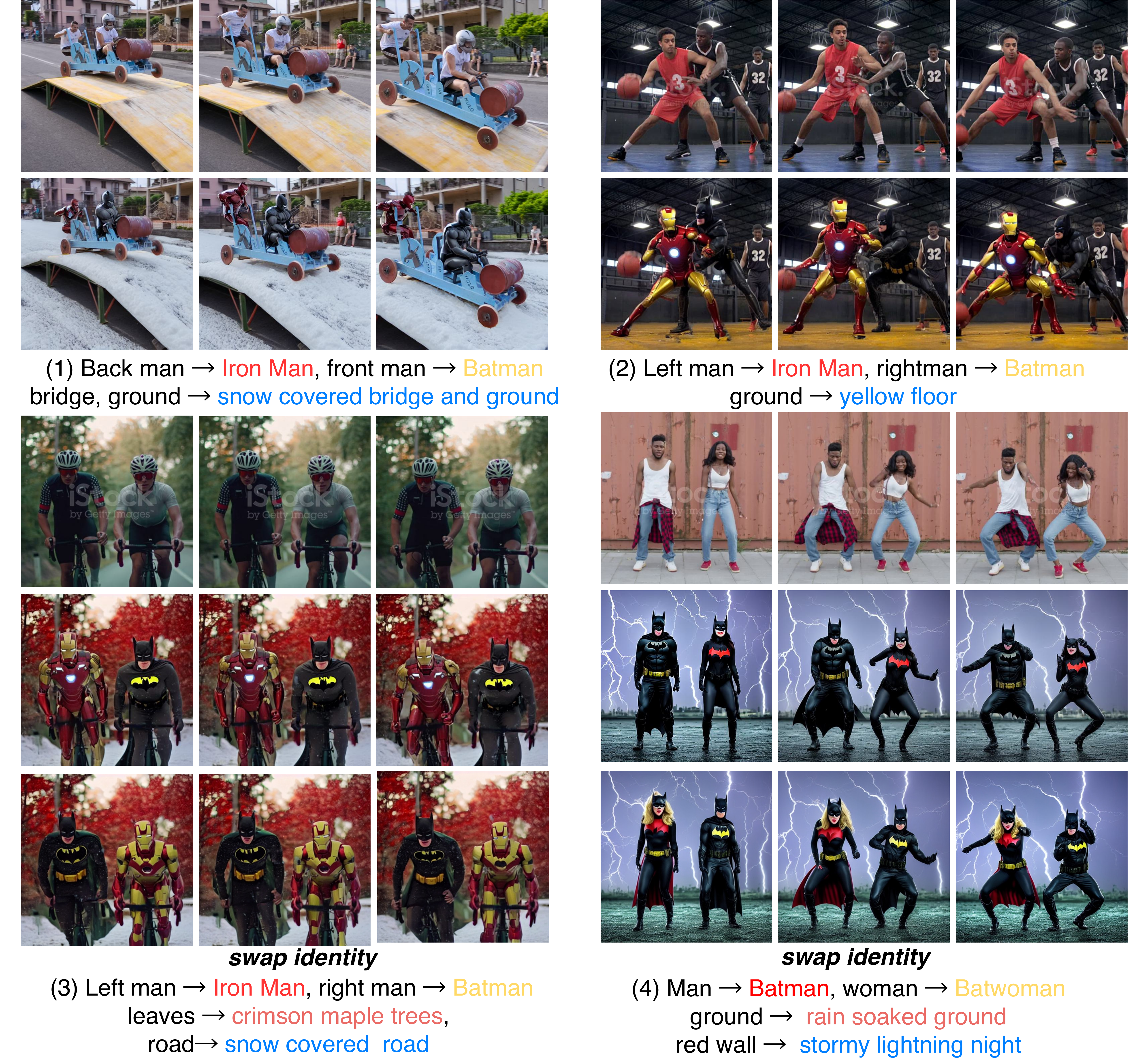}
  \vspace{-6mm}
  \caption{Multi-object multi-attribute editing results. Our EVA supports accurate identity mapping in complex motion videos.}
  \vspace{-8mm}
 \label {eva-multi}
\end{figure*}

\subsection{Results}
\textbf{Single Object Multi-Attribute Editing}
In Fig \ref{eva-single} and \ref{comparison} top, we showcase EVA's editing results in single-object editing. 
Our method maintains the original layout and critical local details like ``railing'' in Fig \ref{eva-single} (1) and ``mountains'' in Fig \ref{comparison} top. 
By decoupling object motion and identity, edited objects seamlessly follow the original movements, even in complex scenarios with view changes.
Additionally, EVA can edit backgrounds that contrast with the original video's style, such as ``a pond under moonlight'' in Fig \ref{eva-single} (3) or ``a stormy lightning night'' in Fig \ref{eva-single} (4).

\noindent\textbf{Multi Object and Attribute Editing}
Fig \ref{eva-multi} displays EVA's multi-object editing outcomes. Our method omits the need for detailed object descriptions. Simple phrases like "a man and another man" suffice to define source objects' identities.
More importantly, our approach enables identity swapping in multi-object scenes, as shown in Fig. \ref{intro} right and Fig. \ref{eva-multi} (3) (4). This is enabled by discrete text embedding control over each attribute.
%since we use discrete text embedding control over each attribute.

\vspace{-5mm}
\subsection{Qualitative and Quantitative Comparisons}

\begin{figure*}[t!]
  \centering
  \includegraphics[width=\linewidth]{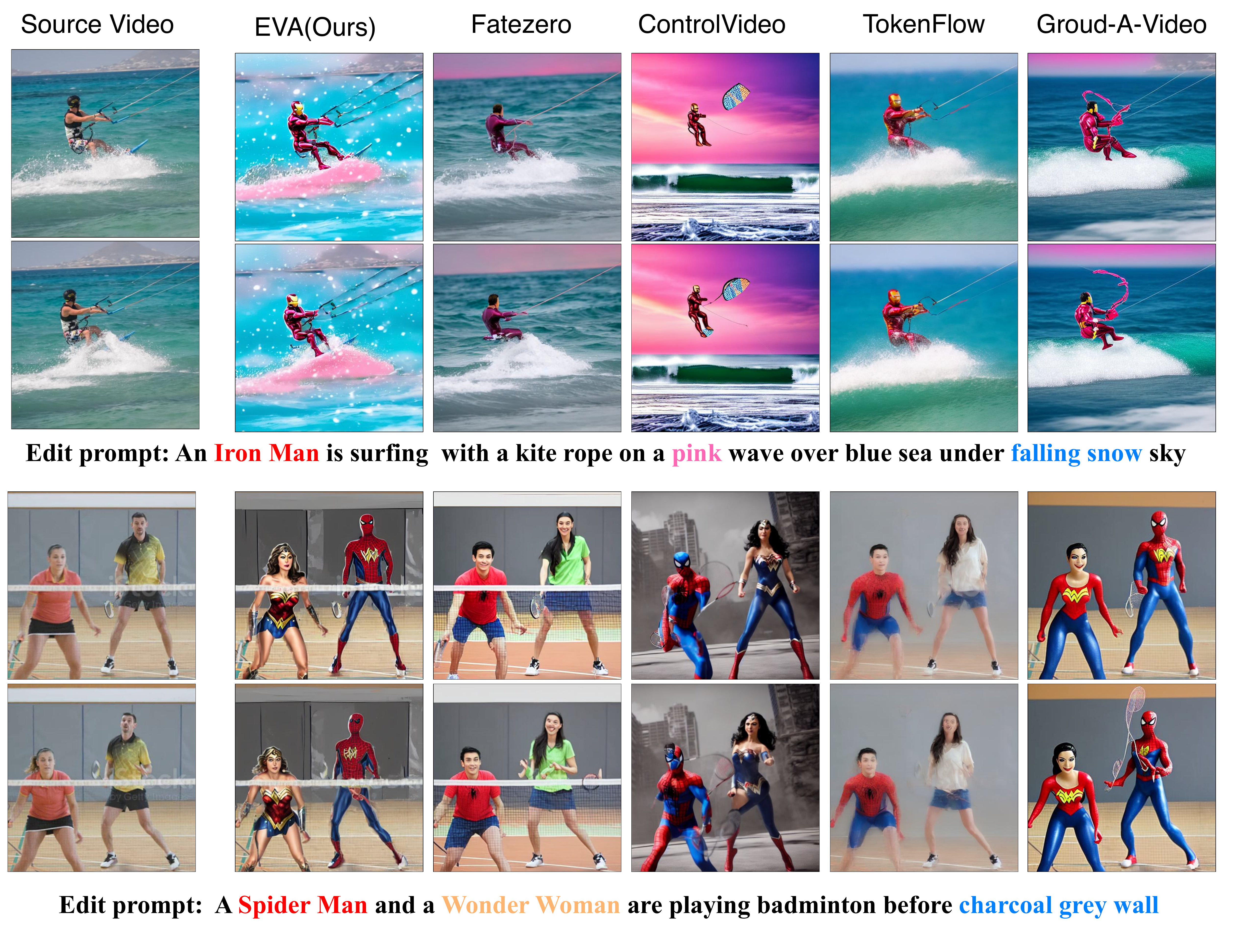}
  \vspace{-8mm}
  \caption{Qualitative comparisons to the existing video editing methods.
  The top figure shows single-object editing results, and the bottom displays multi-object editing results.
  We refer the reader to our \href{https://knightyxp.github.io/EVA/}{project page} for full-video comparisons.}
   \vspace{-8mm}
  \label {comparison}
\end{figure*}

\textbf{Qualitative Comparison}
Fig \ref{comparison} compares our editing results with other baseline methods on single/multi-object videos.
(1). In single-object editing (Fig \ref{comparison} top), FateZero \cite{qi2023fatezero} failed to edit the object and mistakenly edited the pink wave onto the sky. This error arose because attention weights were not precisely aligned with each attribute's words before the word swap.
ControlVideo \cite{zhang2023controlvideo} also incorrectly modified the pink wave onto the sky and could not preserve the layout of the source video. TokenFlow \cite{geyer2023tokenflow} edits the object into ``Ironman" but erases the background mountains and is unable to edit the background. Ground-A-Video \cite{jeong2023ground}, using word-to-bounding box control, confuses the ``kite rope" with the ``Iron Man" and fails to edit the ``snow sky" and ``pink wave''. 
It struggles with preserving local details within the bounding box and lacks awareness of negative examples for adjacent layout weights, which should be mutually exclusive.
(2). In multi-object editing (Fig \ref{comparison} bottom), FateZero, ControlVideo, and Tokenflow all mistakenly confused the ``man" and ``woman" subjects due to a lack of text-to-attribute control. Ground-A-Video edits the ``man" into ``Spiderman" but suffers from attention leakage, where textures of ``Spiderman" leak onto the ``woman," and it fails to retain the local details of the source video, such as ``badminton rackets" and ``nets."
For additional comparison, please refer to the \href{https://knightyxp.github.io/EVA/}{project page}. 

\begin{table}
\centering
\resizebox{1.00\columnwidth}{!}{
\tablestyle{6pt}{1}
\begin{tabular}{c||ccccc}
\hline
\rowcolor{mygray}
Method  &  Frame Acc $\uparrow$ & CLIP-F $\uparrow$ & CLIP-T  $\uparrow$ & Warp-Err $\downarrow$ & $\text{Q}_{edit}$ $\uparrow$  \\
 \hline\hline
FateZero & 73.68 & 95.75 & 33.78  &3.08 &\cellcolor[HTML]{D0F0C0} 10.96 \\
ControlVideo & 95.03 & \textbf{97.71} &  34.41  & 4.73 &\cellcolor[HTML]{D0F0C0} 7.27 \\
TokenFlow & 89.26 & 96.48  & 34.59  & 2.82 &\cellcolor[HTML]{D0F0C0} 12.28 \\
Ground-A-Video & 95.03 & 95.17 & 35.09  & 4.43 &\cellcolor[HTML]{D0F0C0} 7.92\\
\textbf{EVA(ours)}& \textbf{98.92} & 96.09 & \textbf{36.56}  & \textbf{2.73} & \cellcolor[HTML]{D0F0C0} \textbf{13.39} \\
\hline
      \end{tabular}
   }
\captionsetup{font=small}
\caption{\small Quantitative comparison with other methods, the best results are \textbf{bolded}}
\label{table:contrast}
\vspace{-10mm}
\end{table}
\textbf{Quantitative Comparison} \textbf{(1) Automatic Metrics}
Table \ref{table:contrast} presents a quantitative comparison with other methods. Our EVA, with precise text-to-attribute control, achieves the highest frame edit accuracy and the best CLIP-T scores. Although our frame consistency on CLIP-F is slightly lower than ControlVideo \cite{zhang2023controlvideo}, we significantly outperform it in Warp-error and CLIP-T scores. 
This stems from the fact that ControlVideo keeps the background static. Evaluating temporal consistency with pixel-level optical flow provides a more accurate measure than calculating a global temporal score with CLIP \cite{radford2021learning}.
Moreover, our method achieves the highest overall editing score $Q_{edit}$. In general, our EVA demonstrates superior performance on all evaluation metrics.

\textbf{(2)User Study}
While automatic metrics provide a general comparison, they often fail to align well with human perception \cite{liu2023evalcrafter} and cannot accurately verify the accurate editing of each local attribute or the preservation of layout and undesired editing areas. 
Therefore, we conducted a user study for a more detailed comparison. We evaluated the quality of edited videos from four aspects: (1). Subject edit accuracy (accuracy of each attribute's editing), (2). Layout preservation (accuracy of preserving undesired editing areas and overall layout), (3). Motion Alignment, and (4). Overall Preference.

\begin{table}
\vspace{-6mm}
\centering
\small
\resizebox{0.95\columnwidth}{!}{
\tablestyle{4pt}{1}
\begin{tabular}{c||ccccc}
\hline
\rowcolor{mygray}
 &  Subject   &  Layout & Motion   & Overall  \\
\rowcolor{mygray} Method  &   Edit Acc $\uparrow$ &  Preservation $\uparrow$ &  Alignment $\uparrow$ & Preference $\uparrow$  \\
 \hline\hline
FateZero &  2.99 &  3.37 & 3.93 &\cellcolor[HTML]{D0F0C0} 2.98 \\
ControlVideo & 2.66 & 2.04 &  2.50  &\cellcolor[HTML]{D0F0C0} 2.18 \\
TokenFlow & 2.27 & 2.65  & 2.52   &\cellcolor[HTML]{D0F0C0} 1.99 \\
Ground-A-Video & 3.45 & 3.64 & 3.60  &\cellcolor[HTML]{D0F0C0} 3.16\\
\textbf{EVA(ours)}& \textbf{4.42} & \textbf{4.21} & \textbf{4.25}  & \cellcolor[HTML]{D0F0C0} \textbf{4.15} \\
\hline
      \end{tabular}
   }
\captionsetup{font=small}
\caption{\small User study comparison with other methods, The number denotes the average score on a scale from 1 to 5 (worst to best). The best results are \textbf{bolded}.}
\label{table:human evaluation}
\vspace{-8mm}
\end{table}
\vspace{-2mm}

We invited 20 participants to rate 78 video-text pairs on a scale of 1 to 5 across these four criteria. Table \ref{table:human evaluation} shows that our method significantly outperformed FateZero, Tokenflow, and ControlVideo in subject edit accuracy, layout preservation, and motion alignment. Furthermore, our approach exceeded competing related work Ground-A-Video \cite{jeong2023ground} in four human evaluation metrics.
\vspace{-5mm}
\subsection{Ablation Study}
\vspace{-2mm}
To assess the contributions of different components in our proposed EVA framework, we conducted ablation studies with the following designs:

\begin{figure}[t]
  \centering
  \includegraphics[width=\linewidth]{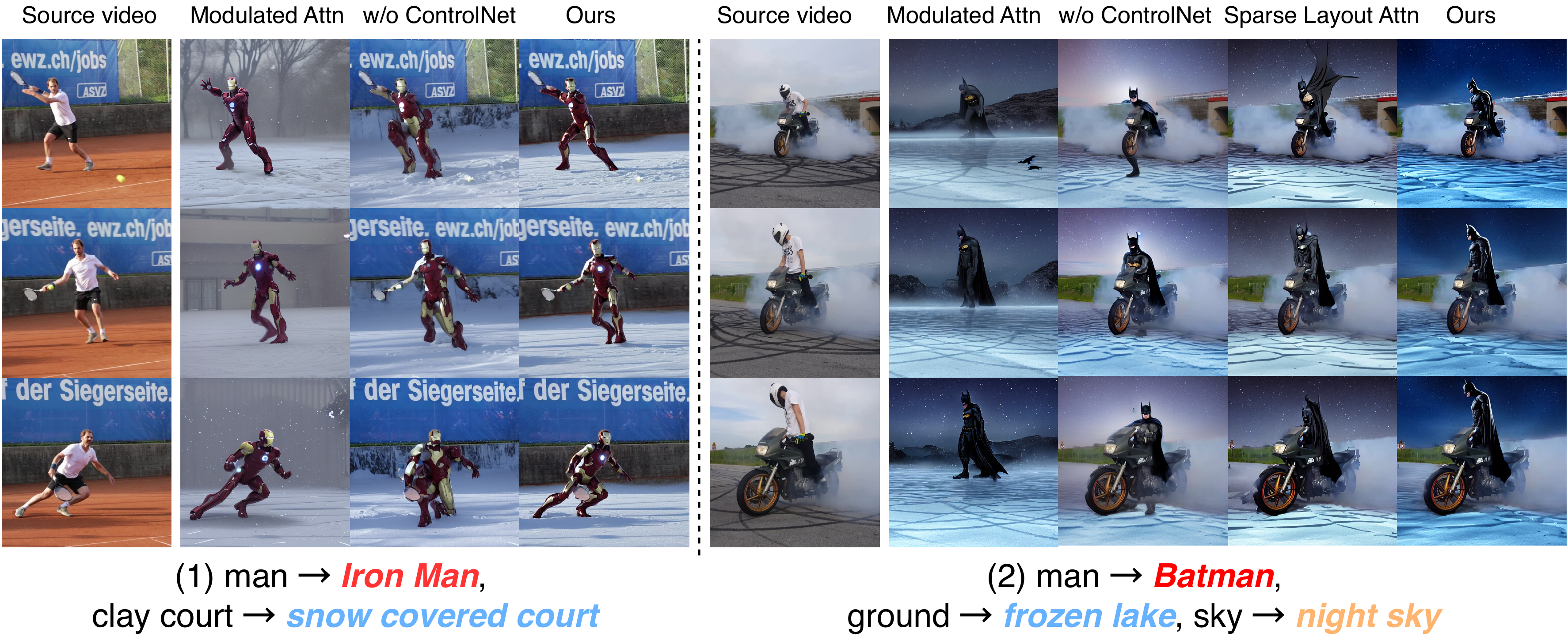}
  \vspace{-7mm}
  \caption{Comparison of Modulated Attention \cite{densediffusion}, the absence of ControlNet, and EVA in complex motion, with Sparse Layout Attn results for video object size changes displayed on the right.}
  % \vspace{-7mm}
  \label {ablation-1}
\end{figure}

\begin{table}
\centering
\small
\resizebox{0.95\columnwidth}{!}{
\tablestyle{4pt}{1}
\begin{tabular}{c||ccccc}
\hline
\rowcolor{mygray}
Method   & CLIP-F $\uparrow$ & CLIP-T  $\uparrow$ & Warp-Err $\downarrow$ & $\text{Q}_{edit}$ $\uparrow$  \\
 \hline\hline
w/o ControlNet & 94.72 & 36.00  &2.75 &\cellcolor[HTML]{D0F0C0} 13.09  \\
w/o Latent Blend  & \textbf{97.02} &  32.56  & 2.88 &\cellcolor[HTML]{D0F0C0} 11.31 \\
w/o Layout guidance& 95.31  & 35.37  & 3.11 &\cellcolor[HTML]{D0F0C0} 11.37 \\
Sparse-casual Layout Attn  & 95.75 & 35.63  & 2.83 &\cellcolor[HTML]{D0F0C0} 12.59\\
\textbf{EVA(ours)}&  96.09 & \textbf{36.56}  & \textbf{2.73} & \cellcolor[HTML]{D0F0C0} \textbf{13.39} \\
\hline
      \end{tabular}
   }
\captionsetup{font=small}
\caption{\small Quantitative ablation of key components of EVA.}
\label{table:ablation}
\vspace{-7mm}
\end{table}
\vspace{-5mm}

\noindent\textbf{Latent Blend} 
In the second and last column of Fig \ref{ablation-1} (1) and (2),
we compare the original modulated attention in DenseDiffusion \cite{densediffusion} with our method.
For fairness, we equip it with ControlNet pose guidance. 
Our findings show that the modulated attention fails to maintain the source background, resulting in varied backgrounds across frames under the same random seed. 
Furthermore, using \cite{densediffusion} alone struggles to preserve details like the ``tennis racket" in Fig \ref{ablation-1} (1) and ``motorcycle" ,``smoke'' in Fig \ref{ablation-1} (2).

\noindent\textbf{ControlNet} 
Next, we ablate the use of ControlNet-Pose, showcased in the third column of Fig \ref{ablation-1}. It is evident that the edited result's posture does not match the source human posture. Therefore, in human-centric complex motion videos, employing pose conditions for intra-object structure information is necessary.

\begin{figure}[t]
  \centering
  \includegraphics[width=\linewidth]{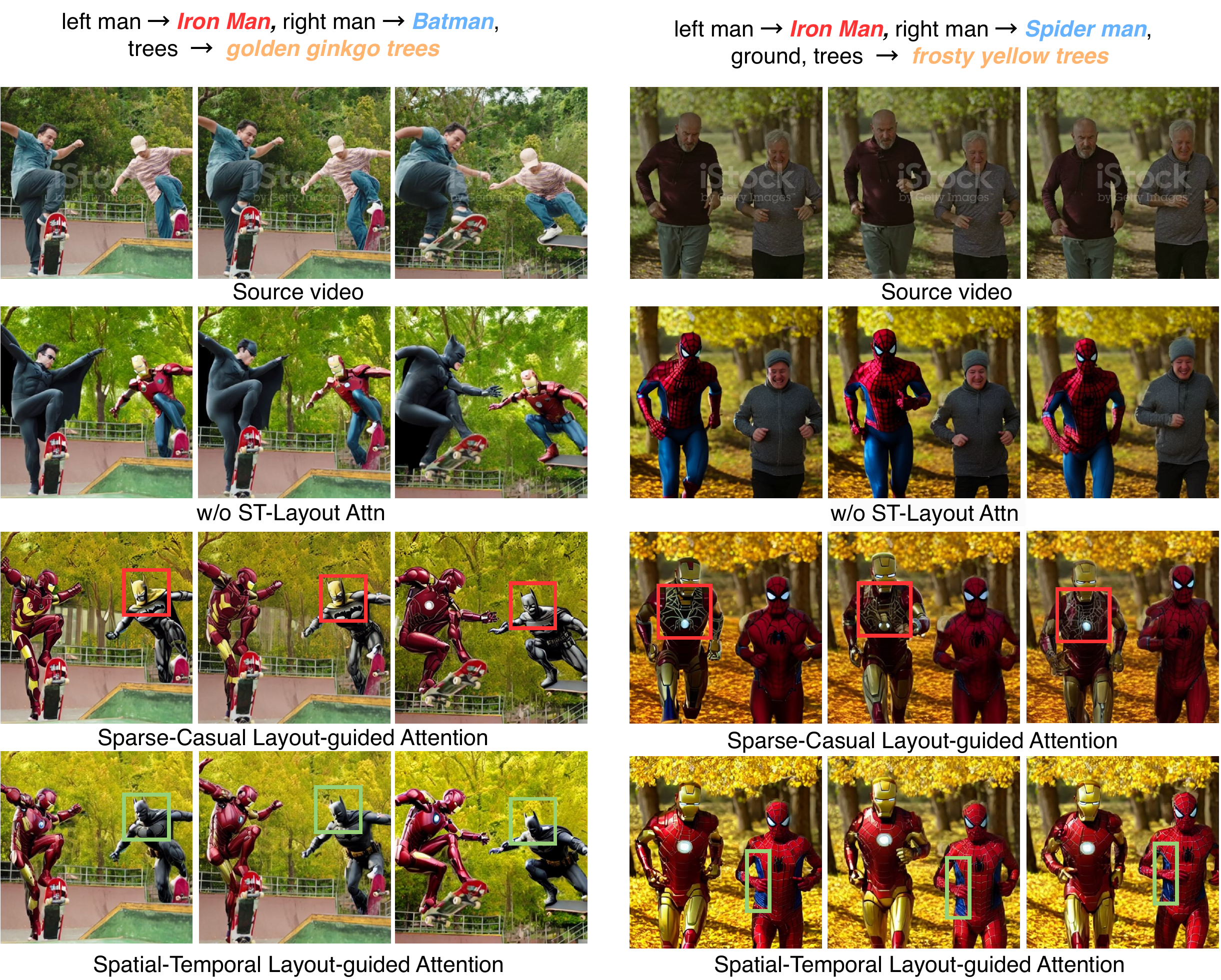}
  \vspace{-6mm}
  \caption{Qualitative comparison of results without ST-Layout Attn, Sparse-Casual Layout-guided Attention (SC-Layout Attn) and our Spatial-Temporal Layout-guided Attention (ST-Layout Attn). Our method results in accurate identity mapping and distinct local details without attention leakage.}
  \vspace{-4mm}
  \label {ablation-st}
\end{figure}

\noindent \textbf{Spatial-Temporal Layout-Guided Attention} 
Fig \ref{ablation-st} contrasts three conditions: without ST-Layout Attn (second row), Sparse Casual Layout-guided Attention (SC-Layout Attn, third row), and with ST-Layout Attn (fourth row). As shown in the second row, the absence of our ST-Layout Attn leads to incorrect identity mapping. For instance, the left man was supposed to be ``Iron Man," and the right is ``Batman," but their identities were swapped in the second row of Fig \ref{ablation-st} left. This underlines the effectiveness of our ST-Layout Attn on accurate identity mapping in multi-object scenes.

Also, sparse layout (the first and the preceding frame) guidance exhibits several limitations, notably: 
(1) Limited Receptive Field for Negative Values: The sparse method's reduced receptive field for query tokens positive/negative value selection across different layouts. The unsuitable selection of negative values results in attention leakage, manifesting as a yellow head of ``Batman'' in Fig \ref{ablation-st} left third row and disordered web-like textures in Fig \ref{ablation-st} right across Iron Man's chest (red box).
(2) Reduced Interaction Across Full Frames: A lack of interaction across the entire video frames results in the loss of local details, such as the distinctive blue sides of Spider-Man (green box in Fig \ref{ablation-st} right). Moreover, this limited interaction contributes to an overall duller color tone.
The quantitative results in Table \ref{table:ablation} further confirm the effectiveness of ST-Layout Attn.
% This sparse interaction across frames leads to disordered textures on Iron Man's chest. In contrast, our full-frame spatial-temporal layout-guided attention not only renders clearer textures for Iron Man and Spider-Man but also enhances detail editing, such as the distinctive blue sides of Spider-Man.

% For additional qualitative comparisons and quantitative results for our ablation studies, please refer to the \href{https://knightyxp.github.io/EVA/}{project page}.
% \vspace{-3mm}

\section{Conclusion}
\vspace{-2mm}
In scenarios of complex human-centric motion, we propose EVA, a general framework for multi-attribute and multi-object video editing. 
We introduce a Spatial-Temporal Layout-Guided Attention mechanism, which leverages the intrinsic positive and negative correspondences of cross-frame diffusion features; 
our ST-Layout Attn not only ensures that each text embedding concentrates on its respective attribute, but also enhances the internal coherence within attributes and keeps the exclusivity of attention weights among different attributes. 
Benefiting from precise attention weighting, EVA can be extended to editing in multi-object scenes. We demonstrate EVA's superior performance in multi-attribute and multi-object editing through extensive experiments.

\bibliographystyle{splncs04}
\bibliography{main}

%%%%%%%%%%%%%%%%%%%%%%%%%%%%%%%%%%%%%%%%%%%%%%%%%%%%%%%%%%%%%%%%%%%%%%%%%%%%%%%
%%%%%%%%%%%%%%%%%%%%%%%%%%%%%%%%%%%%%%%%%%%%%%%%%%%%%%%%%%%%%%%%%%%%%%%%%%%%%%%
% APPENDIX
%%%%%%%%%%%%%%%%%%%%%%%%%%%%%%%%%%%%%%%%%%%%%%%%%%%%%%%%%%%%%%%%%%%%%%%%%%%%%%%
%%%%%%%%%%%%%%%%%%%%%%%%%%%%%%%%%%%%%%%%%%%%%%%%%%%%%%%%%%%%%%%%%%%%%%%%%%%%%%%
% \newpage
% \appendix
% \onecolumn

\end{document}